\pdfoutput=1

\documentclass[11pt]{article}

\usepackage{ACL}

\usepackage{times}
\usepackage{latexsym}

\usepackage[T1]{fontenc}

\usepackage[utf8]{inputenc}

\usepackage{microtype}

\usepackage{inconsolata}

\usepackage{amsthm,amsmath,amssymb}
\usepackage{bbding}
\usepackage{float}
\usepackage{bm}
\usepackage{booktabs}
\usepackage{graphicx}
\usepackage{multirow}
\usepackage{diagbox}
\usepackage{enumitem}
\usepackage{setspace}
\usepackage{algorithm}
\usepackage{algorithmic}

\title{\textsc{SPOR}: A Comprehensive and Practical Evaluation Method for Compositional Generalization in Data-to-Text Generation}

\author{Ziyao Xu, Houfeng Wang \\
  National Key Laboratory for Multimedia Information Processing, Peking University \\
  \texttt{\{xzyxzy,wanghf\}@pku.edu.cn}}

\begin{document}
\maketitle
\begin{abstract}

Compositional generalization is an important ability of language models and has many different manifestations. For data-to-text generation, previous research on this ability is limited to a single manifestation called \textit{Systematicity} and lacks consideration of large language models (LLMs), which cannot fully cover practical application scenarios. In this work, we propose \textsc{SPOR}, a comprehensive and practical evaluation method for compositional generalization in data-to-text generation. \textsc{SPOR} includes four aspects of manifestations (\textit{Systematicity}, \textit{Productivity}, \textit{Order invariance}, and \textit{Rule learnability}) and allows high-quality evaluation without additional manual annotations based on existing datasets. We demonstrate \textsc{SPOR} on two different datasets and evaluate some existing language models including LLMs. We find that the models are deficient in various aspects of the evaluation and need further improvement. Our work shows the necessity for comprehensive research on different manifestations of compositional generalization in data-to-text generation and provides a framework for evaluation.  The dataset and code are available at \url{https://github.com/xzy-xzy/SPOR}.

\end{abstract}

\section{Introduction}

\par Data-to-text generation \cite{gen-survey} is an important task in natural language generation (NLG). It aims to generate fluent and faithful text based on structured data input and is critical in many NLG systems, such as report generation \cite{rotowire}, oriented dialogues \cite{Self}, etc. In data-to-text generation, structured data input is compositional, i.e., it can be considered as a combination of elements formed according to certain rules. Therefore, in order to handle the practical data-to-text generation, the language models should have the ability to recombine previously learned elements with certain rules to map new inputs made up from these elements to their correct output \cite{summary}, which is the so-called compositional generalization.
\par Compositional generalization is an important ability of language models for many tasks. In semantic parsing and mathematical reasoning tasks, many different manifestations of this ability have been studied \citep{PCFG, Trans}, such as \textit{systematicity} (handle combinations unseen during training), \textit{productivity} (extrapolate to longer sequences than those seen during training), etc. For compositional generalization in data-to-text generation, only \textit{systematicity} receives attention \cite{Self}, and research on other manifestations is lacking. The single \textit{systematic} manifestation cannot fully cover practical application scenarios of compositional generalization and cannot comprehensively reflect this ability of language models in data-to-text generation. Although research on different manifestations of compositional generalization in data-to-text generation is necessary, there is currently no comprehensive evaluation method to support such research.

\par To solve this problem, we propose \textsc{SPOR}, a comprehensive and practical evaluation method for compositional generalization in a data-to-text generation. Based on the manifestations of compositional generalization mentioned in \citet{PCFG}, \textsc{SPOR} includes four aspects of compositional generalization in data-to-text generation:

\begin{itemize}[itemsep=2pt,topsep=2pt,parsep=2pt]

\item[$\bullet$]\textit{Systematicity.} The ability to handle data combinations unseen during training.

\item[$\bullet$]\textit{Productivity.} The ability to handle a larger amount of data within a sample than seen during training.

\item[$\bullet$]\textit{Order invariance.} The ability to maintain the fidelity and proper data ordering of the output text when the input order of data in an unordered set is changed.

\item[$\bullet$]\textit{Rule learnability.} The ability to actually learn and apply \textit{copy rule} for generation, rather than memorize specific mappings.

\end{itemize}

\par For each aspect, we propose the corresponding methods for dataset construction and evaluation. Based on existing datasets, we mainly perform repartition \cite{CFQ} and element modification to construct datasets for our evaluation. Overall, the evaluation method \textsc{SPOR} has the following properties:

\begin{itemize}[itemsep=2pt,topsep=2pt,parsep=2pt]

\item[$\bullet$]\textit{Necessity.} The ability or property in each aspect manifests compositional generalization and is required by the model for practical data-to-text generation.

\item[$\bullet$]\textit{High evaluation quality.} For each aspect, the evaluation method can effectively evaluate the corresponding ability or property.

\item[$\bullet$]\textit{Low construction cost.} Based on existing datasets, the dataset used for evaluation does not require additional manual annotation and can be constructed automatically. 

\end{itemize}

\par We demonstrate \textsc{SPOR} on two existing datasets for data-to-text generation and evaluate some existing language models. Previous research on compositional generalization in data-to-text generation lacks consideration of large language models (LLMs) due to the lack of methods to directly fine-tune and apply LLMs to data-to-text generation in the past. Nowadays, advanced Parameter-Efficient Fine-Tuning such as LoRA \cite{LoRA} provides the methods, and the consideration of LLMs becomes necessary. Therefore, we include some advanced LLMs in our evaluation to partially fill the gap in previous research.
\section{Preliminaries}
In this section, we provide a brief description of the datasets that \textsc{SPOR} is demonstrated on, the evaluated models, and the evaluation metrics.

\subsection{Datasets}
\par
\begin{figure}
    \centering
    \includegraphics[width=\columnwidth]{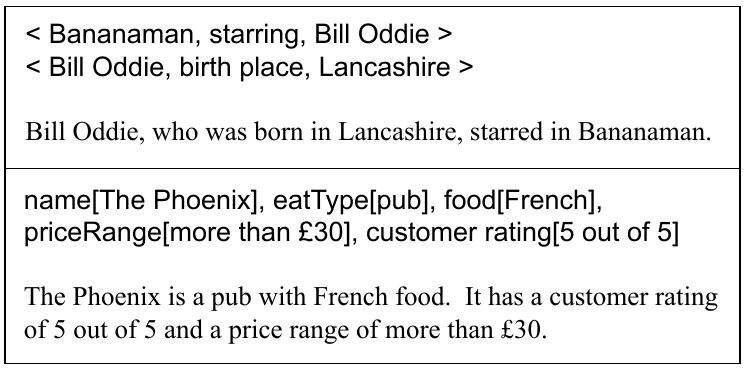}
    \caption{Examples of data-text pairs in WebNLG (above) and E2E (below).}
    \label{fig:dataset}
\end{figure}

 We demonstrate \textsc{SPOR} on two data-to-text generation datasets, WebNLG \cite{WebNLG} and E2E \cite{E2E}. Both contain $(D,T)$ pairs, where $D$ is the input data and $T$ is the text that verbalizes the data. Figure~\ref{fig:dataset} shows examples of data-text pairs in WebNLG and E2E.
\par WebNLG is a realistic multi-domain dataset. In WebNLG, $D$ is an unordered set of 1\textasciitilde7 triples $\langle s, p, o \rangle $, where $s,p,o$ represents subject, predicate, and object, respectively. We regard triples as data units for WebNLG. In the original WebNLG dataset, 10 domains are present in the training set and can be used in the evaluation. We select the latest version, WebNLG+ \cite{WebNLG+}, which increases the number of available domains to 16 and contains more samples. For the samples used for testing, we retain only samples in which all data units appear in the training set. After processing, WebNLG+ contains 3,873 distinct triples, 13,211 samples in the training set, and 2,179 samples in the test set.
\par E2E is a dataset in the restaurant domain. In E2E, $D$ is a name with an unordered set of 1\textasciitilde7 pairs $(a, v)$, where $a,v$ represents attribute and value, respectively. We regard attribute-value pairs as data units for E2E. We select the cleaned version \cite{E2EC}, which fixes the data to eliminate inconsistencies between the data and the text. We perform further filtering based on the clean version, retaining only samples in which all input values have matches in the text. After processing, E2E contains 7 distinct attributes, 45 distinct attribute-value pairs, 6,735 samples in the training set, and 1,635 samples in the test set.

\subsection{Models}
We evaluate some smaller-sized, previously state-of-the-art language models in data-to-text generation, including two encoder-decoder language models T5-large \cite{T5} and BART-large \cite{BART}, and one causal language model GPT-2-large \cite{GPT2}. We also evaluate some advanced LLMs, including one encoder-decoder language model T5-11b \cite{t5-11b}, and two causal language models Mistral-7b \cite{Mistral} and Llama-2-13b \cite{llama2}. For data input, we use the linearization method \cite{T5forD2T}. Following previous work in data-to-text generation \cite{Self}, we use fine-tuning method and treat the fine-tuning phase as the training phase. We use LoRA fine-tuning, which has better performance than full fine-tuning in data-to-text generation \cite{LoRA}. For model training, the optimizer is Adam \cite{adam}. The learning rate is 1e-4, and the batch size is 6. For the LoRA setting, we use $r=8$, $a=32$, and 0.1 dropout. We train the models for 10 epochs. For model inference, the beam width is 5. See Appendix~\ref{app:model} for more details about model size, input, training, and inference.
\subsection{Metrics}
\label{sec:metric}
\par We use PARENT \cite{PARENT} as the performance metric to measure the quality of the model's output. PARENT is a metric designed for data-to-text generation tasks, which considers the alignment of the output to both input data and reference texts. PARENT better reflects the semantic fidelity of the output and has a stronger correlation with human judgments than reference-only-based metrics. Metrics other than the performance metric are described in the corresponding aspects.

\section{Evaluation Method}
In this section, we describe each aspect of \textsc{SPOR}. Each subsection corresponds to an aspect that includes: (1) the overview; (2) how to construct the dataset; (3) the statistics of the dataset; (4) how to perform the evaluation and (5) the results and analysis. For all results reported, we run experiments three times with different random seeds and average the results to avoid contingency. Appendix~\ref{app:qua} provides the qualitative analysis of evaluations, showing specific samples with model outputs.

\subsection{Systematicity}

\begin{figure}
    \centering
    \includegraphics[width=\columnwidth]{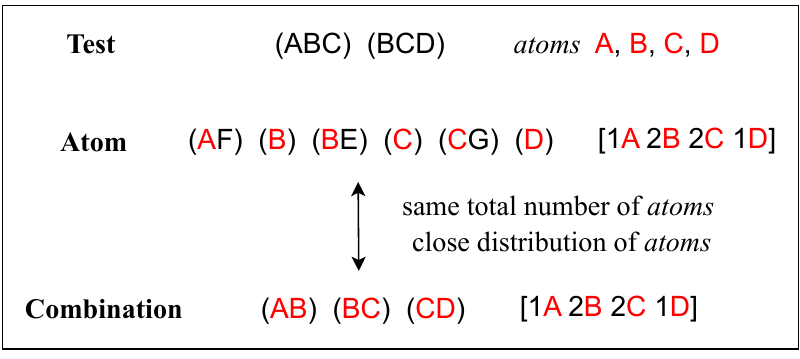}
    \caption{An example of datasets for the \textit{systematicity} evaluation. Each pair of brackets denotes a sample and each letter (A\textasciitilde G) denotes a data unit.}
    \label{fig:sys}
\end{figure}

The first aspect we evaluate is \textit{systematicity} \cite{PCFG}. \textit{Systematicity} is a notion frequently used in tests of compositional generalization \citep{SCAN, COGS, PCFG, CFQ}, which refers to the ability to handle combinations of known elements that are not seen during training. In the data-to-text generation task, the elements refer to the data. Although a large corpus allows the model to see a large amount of data, the possible combinations of data are too numerous to be fully covered. In practical applications, the model will often see combinations of known data in the input that are not seen during training, so the ability to handle unseen combinations of data is important. 

In the \textit{systematicity} evaluation, by reconstructing the dataset, we allow the model to see all data units in the test set during training, but not any combination of them. In this case, the model needs \textit{systematicity} to handle unseen combinations at test time. We use the model performance in this case as the \textit{systematicity} metric. Based on the same test set, we also construct the case where the model can see combinations of data units to test whether the model's performance when it cannot see combinations is comparable to that when it can.

\subsubsection{Dataset Construction}
\label{sec:main_ent}
We construct one test set and two training sets \textbf{Atom (A)} and \textbf{Combination (C)}. Figure~\ref{fig:sys} illustrates the goal of our construction. We call the data units that appear in the test set \textit{atoms}. Both \textbf{Atom} and \textbf{Combination} cover all \textit{atoms}, and they have the same total number of \textit{atoms} and close distribution of \textit{atoms}. However, \textbf{Atom} does not contain any combination of \textit{atoms}, but \textbf{Combination} does.

\begin{algorithm}[t]
\caption{Construction of \textbf{Atom} and the test set}
\begin{algorithmic}
\label{algo:sys1}
\renewcommand{\algorithmicrequire}{\textbf{Input:}}
\renewcommand{\algorithmicensure}{\textbf{Output:}}
\small
\setstretch{1.1}
\REQUIRE{original dataset $S$}
\ENSURE{\textbf{Atom} ($A$), test set ($T$), \textbf{Blocked} ($B$)}

\STATE $T, A, B \leftarrow \varnothing$
\WHILE{$S \not= \varnothing$}
{
    \STATE $x \leftarrow $ randomly selected sample in $S$
    \STATE $S \leftarrow S - \{x\}$ 
    \STATE $R \leftarrow \{y \ | \ y \in A \cup S \ \land \ y \notin B \ \land \ |y \cap x|=1\}$
    \IF{$x \subseteq \bigcup{R} \ \text{ and } \max_{y \in A}{|y \cap x|} \leq 1 $}
    {
        \STATE $T \leftarrow T \cup \{x\}$
        \STATE $S \leftarrow S - R$
        \STATE $A \leftarrow A \cup R$
        \STATE $B \leftarrow B \cup \{y \ | \ y \in S \ \land \ |y \cap x|>1\}$
    }
    \ENDIF
}
\ENDWHILE
\end{algorithmic}
\end{algorithm}

\par We use Algorithm~\ref{algo:sys1} to construct \textbf{Atom} and the test set. We assume that the original dataset is the set $S$ and each sample $x$ in $S$ is a set of data units. For a set $x$, we use $|x|$ to denote the number of data units it contains. For a set $S$ containing sets, we use $\bigcup S$ to denote the union of the sets it contains, i.e., $\bigcup S$ is the set of all data units occurring in $S$. 
\par Initially, both \textbf{Atom} and the test set are empty sets, and we set an initially empty auxiliary set \textbf{Blocked} to store samples containing combinations of \textit{atoms}. Each time, we remove a sample $x$ from $S$ and check all samples in the current \textbf{Atom} and samples in $S$ that are not in \textbf{Blocked} and include only one data unit in $x$. If these samples cover all data units in $x$, and \textbf{Atom} does not contain combinations of data units in $x$, then we:
\begin{itemize}[itemsep=2pt,topsep=2pt,parsep=2pt]

\item[$\bullet$]Add $x$ to the test set.

\item[$\bullet$]Remove samples in $S$ that are not in \textbf{Blocked} and include only one data unit in $x$, and add them to \textbf{Atom}.

\item[$\bullet$]Add samples in $S$ that include more than one data unit in $x$ to \textbf{Blocked}.
\end{itemize}

\par This process is repeated until $S$ is empty. Under this construction method, \textbf{Atom} covers all \textit{atoms} but does not contain any combination of \textit{atoms}. The samples containing combinations of \textit{atoms} are all in \textbf{Blocked}. 

\begin{algorithm}[t]
\caption{Construction of \textbf{Combination}}
\begin{algorithmic}
\label{algo:sys2}
\renewcommand{\algorithmicrequire}{\textbf{Input:}}
\renewcommand{\algorithmicensure}{\textbf{Output:}}
\small
\setstretch{1.1}
\REQUIRE{\textbf{Atom} ($A$), test set ($T$), \textbf{Blocked} ($B$), divergence measure function $\mathcal D$, threshold $r$}
\ENSURE{\textbf{Combination} ($C$)}

\STATE $C, A'\leftarrow A$
\STATE $B\leftarrow B - T$
\STATE $T\leftarrow \bigcup T$
\STATE \textbf{define function} $\mathcal{F}(x, G)$ \textbf{as} $\sum_{y \in G}{|x \cap y|}$
\STATE \textbf{define function} $\mathcal{V}(x)$ \textbf{as} $\mathcal{F}(x\cap T, A) - \mathcal{F}(x\cap T, C)$
\WHILE{$B \not= \varnothing$}
{
    \STATE $x \leftarrow $ sample in $S$ with maximum $\mathcal{V}(x)$
    \STATE $B \leftarrow B - \{x\}$ 
    \STATE $R \leftarrow \varnothing $
    \FORALL {$y \in A'$ in ascending order of $\mathcal{V}(y)$}
    {
        \STATE $R' \leftarrow R \cup \{y\}$
        \IF {$|\bigcup R'|\leq|x|$ \AND $T \subseteq (C-R')\cup\{x\}$}
        {
            \STATE $R \leftarrow R'$
        }
        \ENDIF
    }
    \ENDFOR
    \IF {$|\bigcup R|=|x|$ \AND $\mathcal D(A, (C-R)\cup\{x\}) \leq r$}
    {
        \STATE $C \leftarrow (C-R)\cup\{x\}$
        \STATE $A' \leftarrow A'-R$
    }
    \ENDIF
}
\ENDWHILE
\end{algorithmic}
\end{algorithm}    
\par We then use Algorithm~\ref{algo:sys2} to construct \textbf{Combination}. The core idea of Algorithm~\ref{algo:sys2} is to replace samples in \textbf{Atom} with samples that have combinations of \textit{atoms} to obtain \textbf{Combination}. We initialize \textbf{Combination} with \textbf{Atom}. For each sample $x$ in \textbf{Blocked} but not in the test set, we try to replace a cluster of samples belonging to \textbf{Atom} with $x$ in \textbf{Combination}, ensuring that \textbf{Combination} still covers all \textit{atoms} and the total number of \textit{atoms} remains the same after the replacement. Each replacement makes \textbf{Combination} have one more sample with combinations of \textit{atoms}.
\par To ensure that the distributions of \textit{atoms} in \textbf{Atom} and \textbf{Combination} are close, we perform the replacement only if the divergence of the two distributions after the replacement does not exceed a threshold $r$. Following \citet{CFQ}, we measure the divergence using the Chernoff coefficient $\mathcal{D}(P, Q) = 1-\sum_{k}{p_k^{0.5}q_k^{0.5}} \in [0,1] $ \cite{chernoff} 
and set the threshold $r = 0.02$, where $p_k$ and $q_k$ denote the proportion of the atom $k$ in datasets $P$ and $Q$, respectively. Random replacements will cause the divergence to reach the threshold too early. To avoid this, we define $\mathcal{V}(x)$ as the subtraction of the total occurrences of \textit{atoms} from $x$ in \textbf{Atom} and \textbf{Combination}, and try to use samples with high $V(x)$ to replace samples with low $V(x)$. This replacement method controls the growth of divergence, allowing more replacements to occur and thus allowing \textbf{Combination} to contain more combinations of \textit{atoms}.

\begin{table}
\centering

\begin{tabular}{l|cc|cc}
\hline
    & \multicolumn{2}{c|}{WebNLG} & \multicolumn{2}{c}{E2E}\\
\cline{2-5}
    & \textbf{A} & \textbf{C} & \textbf{A} & \textbf{C} \\
\hline
\# samples & 4,717 & 3,256 & 3,351 & 1,390 \\
\# data units & 9,636 & 8,267 & 13,311 & 7,043\\
\# \textit{atoms} & 5,281 & 5,281 & 3,298 & 3,298 \\
\# \textit{pairs} & 0 & 1,969 & 0 & 2,670 \\
\hline

\end{tabular}

\caption{\label{sys_info}Some statistics about the training sets for the \textit{systematicity} evaluation. \textit{Pairs} refer to pairs of \textit{atoms} that co-occur in a sample.}
\end{table}
\subsubsection{Dataset Statistics}
Table~\ref{sys_info} shows the statistics about the training sets for the \textit{systematicity} evaluation. The size of the test set for the \textit{systematicity} evaluation is related to the number of distinct data units contained in the original dataset. For a dataset like E2E with a small number of distinct data units, it is more difficult to construct a large test set. To maximize the size of the test set, we randomly pick $x$ among those with the largest $|x|$ in Algorithm~\ref{algo:sys1}. We perform multiple random constructions and use the one with the largest test set size. The test set contains 2,360 samples on WebNLG and 156 samples on E2E.

\subsubsection{Evaluation}
We train the model on \textbf{Atom} and \textbf{Combination} respectively and test the performance of the two trained models on the test set. We evaluate \textit{systematicity} of the model by the performance on \textbf{Atom}. We use the performance on \textbf{Combination} as a bound to analyze the \textit{systematicity} level of the model.

\begin{table}
\centering
\begin{tabular}{l|p{0.9cm}p{0.9cm}|p{0.9cm}p{0.9cm}}
\hline
            & \multicolumn{2}{c|}{WebNLG}     & \multicolumn{2}{c}{E2E}         \\ \cline{2-5} 
            & \multicolumn{1}{c}{\textbf{A}}     & \multicolumn{1}{c|}{\textbf{C}}    & \multicolumn{1}{c}{\textbf{A}}     & \multicolumn{1}{c}{\textbf{C}}     \\ \hline
T5-large    & 66.14$^\dag$        & 66.54          & 49.19$^\ddag$          & 52.76          \\
BART-large  & 64.44$^\dag$          & 64.80          & 50.49$^\ddag$          & 52.63          \\
GPT-2-large & 63.98$^\ddag$         & 64.93          & 51.82$^\ddag$          & 52.95          \\
T5-11b      & \textbf{68.93} & \textbf{69.07} & \textbf{53.78}$^\ddag$ & \textbf{54.72} \\
Mistral-7b  & 66.87$^\dag$          & 67.09          & 53.06$^\dag$          & 54.22          \\
Llama-2-13b & 65.87$^\dag$          & 66.18          & 51.28$^\ddag$          & 53.35          \\ \hline
\end{tabular}

\caption{\label{sys_res}Performance of models on the two training sets for the \textit{systematicity} evaluation. Significance tests are conducted to check whether the performance of the model on \textbf{Atom} is significantly lower than that on \textbf{Combination}. $\dag$ means $p<0.1$ and $\ddag$ means $p<0.05$.}
\end{table}

\subsubsection{Results and Analysis} 

\par Table~\ref{sys_res} shows the results of the \textit{systematicity} evaluation. On WebNLG, T5-11b performs best on \textbf{Atom}, showing the strongest \textit{systematicity}. Among the LLMs, both T5-11b and Mistral-7b outperform all the smaller LMs on \textbf{Atom}, reflecting an improvement in \textit{systematicity}. However, all models, including LLMs, show performance gaps on \textbf{Atom} and \textbf{Combination}. As \textbf{Atom} and \textbf{Combination} have the same total number of \textit{atoms} and close distribution of \textit{atoms}, the gaps are attributed to differences in the visibility of combinations of \textit{atoms}, indicating that when the model cannot see combinations of \textit{atoms} during training, it is unable to handle combinations of \textit{atoms} as well as when it can see. This reflects a deficiency in \textit{systematicity} of the model. The results on E2E are similar, and the performance gaps on \textbf{Atom} and \textbf{Combination} on E2E are more significant than on WebNLG, which further confirms the deficiency in \textit{systematicity} of the model. In conclusion, the LLMs overall show an improvement in \textit{systematicity} compared to the smaller LMs but do not eliminate the deficiency in \textit{systematicity} of the model.

\subsection{Productivity}
The second aspect we evaluate is \textit{productivity} \cite{PCFG}. \textit{Productivity}, in the context of compositionality, refers to the ability to extrapolate to longer sequences than those seen during training \cite{Trans}.
Similar to \textit{systematicity}, \textit{productivity} is also a notion frequently used in tests of compositional generalization \citep{SCAN, PCFG, Trans}. In the data-to-text generation task, \textit{productivity} corresponds to the ability to handle a larger amount of data in the input than those seen during training. In practical applications, the amount of data contained in an input can be arbitrarily large, and it is impossible for a finite corpus to cover inputs with arbitrarily large amounts of data. The model will often encounter inputs with a larger amount of data than those seen during training and should have the ability to handle this situation.

\par In the \textit{productivity} evaluation, we limit the number of data units of each sample during training, and test how the model performs when handling a larger amount of input data units than those seen during training. On the same test set, we also test the model trained with samples without the limit on the number of input data to see whether the model's performance with the limit is comparable to that without the limit.

\begin{figure}
    \centering
    \includegraphics[width=\columnwidth]{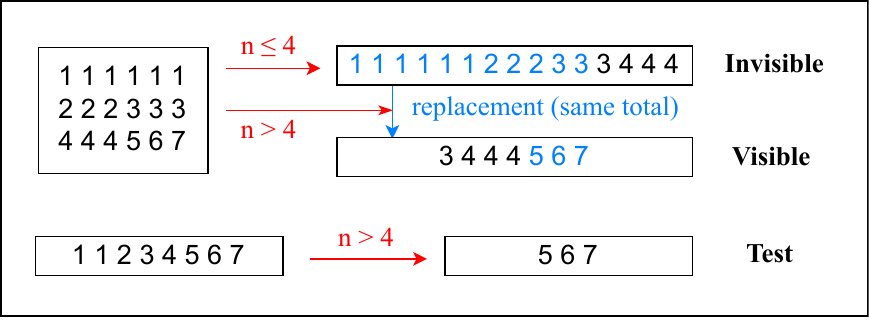}
    \caption{An example of datasets with threshold $N=4$ for the \textit{productivity} evaluation. Each number represents a sample with a corresponding number of data units.}
    \label{fig:pro}
\end{figure}
\begin{table}
\centering
\small
\resizebox{\columnwidth}{!}{
\begin{tabular}{c|c|ccccccc}
\hline
                       &            & \textbf{1} & \textbf{2} & \textbf{3} & \textbf{4} & \textbf{5} & \textbf{6} & \textbf{7} \\ \hline
\multirow{2}{*}{$N=3$} & \textbf{I} & 249        & 193        & 239        & 0          & 0          & 0          & 0          \\
                       & \textbf{V} & 19         & 18         & 9          & 56         & 57         & 44         & 71         \\ \hline
\multirow{2}{*}{$N=4$} & \textbf{I} & 249        & 193        & 239        & 260        & 0          & 0          & 0          \\
                       & \textbf{V} & 0          & 17         & 35         & 25         & 148        & 99         & 117        \\ \hline
\multirow{2}{*}{$N=5$} & \textbf{I} & 249        & 193        & 239        & 260        & 227        & 0          & 0          \\
                       & \textbf{V} & 9          & 52         & 128        & 99         & 34         & 203        & 178        \\ \hline
\multirow{2}{*}{$N=3$} & \textbf{I} & 86         & 592        & 1,480      & 0          & 0          & 0          & 0          \\
                       & \textbf{V} & 0          & 66         & 633        & 0          & 414        & 148        & 103        \\ \hline
\multirow{2}{*}{$N=4$} & \textbf{I} & 86         & 592        & 1,480      & 2,151      & 0          & 0          & 0          \\
                       & \textbf{V} & 0          & 80         & 1,227      & 1,601      & 4          & 543        & 113        \\ \hline
\multirow{2}{*}{$N=5$} & \textbf{I} & 86         & 592        & 1,480      & 2,151      & 1,612      & 0          & 0          \\
                       & \textbf{V} & 0          & 389        & 1,400      & 2,029      & 1,435      & 219        & 113        \\ \hline
\end{tabular}}

\caption{\label{pro_info}Number of samples in training sets for the \textit{productivity} evaluation with each number (from 1 to 7) of data units in WebNLG (above) and E2E (below).}
\vspace{-0.4cm}
\end{table}
\begin{table*}[t]
\centering
\resizebox{\textwidth}{!}{
\begin{tabular}{l|p{0.9cm}p{0.9cm}p{0.9cm}p{0.9cm}p{0.9cm}p{0.9cm}|p{0.9cm}p{0.9cm}p{0.9cm}p{0.9cm}p{0.9cm}p{0.9cm}}
\hline
            & \multicolumn{6}{c|}{WebNLG}                                                                                                                   & \multicolumn{6}{c}{E2E}                                                                                                                       \\ \cline{2-13} 
            & \multicolumn{2}{c|}{$N=3$}                           & \multicolumn{2}{c|}{$N=4$}                           & \multicolumn{2}{c|}{$N=5$}      & \multicolumn{2}{c|}{$N=3$}                           & \multicolumn{2}{c|}{$N=4$}                           & \multicolumn{2}{c}{$N=5$}       \\ \cline{2-13} 
            & \multicolumn{1}{c}{\textbf{I}}      & \multicolumn{1}{c|}{\textbf{V}}     & \multicolumn{1}{c}{\textbf{I}}      & \multicolumn{1}{c|}{\textbf{V}}     & \multicolumn{1}{c}{\textbf{I}}      & \multicolumn{1}{c|}{\textbf{V}}     & \multicolumn{1}{c}{\textbf{I}}      & \multicolumn{1}{c|}{\textbf{V}}     & \multicolumn{1}{c}{\textbf{I}}      & \multicolumn{1}{c|}{\textbf{V}}     & \multicolumn{1}{c}{\textbf{I}}      & \multicolumn{1}{c}{\textbf{V}}          \\ \hline
T5-large    & 68.24$^\ddag$          & \multicolumn{1}{c|}{69.82}          & 68.32$^\ddag$          & \multicolumn{1}{c|}{70.11}          & 68.36$^\dag$          & 68.71          & 61.27$^\ddag$          & \multicolumn{1}{c|}{62.91}          & 64.31$^\ddag$          & \multicolumn{1}{c|}{64.91}          & 63.81          & 64.11          \\
BART-large  & 67.58$^\dag$          & \multicolumn{1}{c|}{69.17}          & 67.54$^\dag$          & \multicolumn{1}{c|}{69.89}          & 68.84          & 69.17          & 62.59          & \multicolumn{1}{c|}{62.98}          & 64.31          & \multicolumn{1}{c|}{64.68}          & 63.37$^\dag$          & 63.71          \\
GPT-2-large & 63.95$^\ddag$          & \multicolumn{1}{c|}{66.43}          & 64.96$^\ddag$          & \multicolumn{1}{c|}{68.61}          & 65.25$^\dag$          & 66.90          & 57.81$^\ddag$          & \multicolumn{1}{c|}{62.89}          & 64.22$^\dag$          & \multicolumn{1}{c|}{65.17}          & 63.99          & 64.15          \\
T5-11b      & \textbf{70.86}$^\ddag$ & \multicolumn{1}{c|}{\textbf{71.10}} & \textbf{70.03} & \multicolumn{1}{c|}{70.15}          & \textbf{69.57}$^\ddag$ & \textbf{69.83} & \textbf{62.79}$^\dag$ & \multicolumn{1}{c|}{63.33}          & 63.97          & \multicolumn{1}{c|}{64.48}          & 63.89$^\dag$          & 64.25          \\
Mistral-7b  & 68.92$^\ddag$          & \multicolumn{1}{c|}{70.55}          & 69.43$^\dag$          & \multicolumn{1}{c|}{\textbf{71.09}} & 69.41          & 69.63          & 62.71$^\ddag$          & \multicolumn{1}{c|}{\textbf{64.53}} & \textbf{65.13}$^\ddag$ & \multicolumn{1}{c|}{\textbf{66.06}} & 64.18          & \textbf{64.82} \\
Llama-2-13b & 68.77$^\ddag$          & \multicolumn{1}{c|}{69.78}          & 69.55$^\dag$          & \multicolumn{1}{c|}{70.30}          & 69.08          & 69.23          & 61.18$^\ddag$          & \multicolumn{1}{c|}{62.76}          & 64.46          & \multicolumn{1}{c|}{64.86}          & \textbf{64.22} & 64.40          \\ \hline
\end{tabular}
}

\caption{\label{pro_res}Performance of models trained on the two training sets with the number threshold $N \in \{3, 4, 5\}$ for the \textit{productivity} evaluation. Significance tests are conducted to check whether the performance of the model on \textbf{Invisible} is significantly lower than that on \textbf{Visible}. $\dag$ means $p<0.1$ and $\ddag$ means $p<0.05$.}
\end{table*}

\subsubsection{Dataset Construction}
We construct one test set and two training sets \textbf{Invisible (I)} and \textbf{Visible (V)}. We start by setting a number threshold $N$. We construct \textbf{Invisible} using all samples with no more than $N$ data units. Similar to Algorithm~\ref{algo:sys2}, we replace the samples in \textbf{Invisible} with samples with more than $N$ data units to obtain \textbf{Visible}, ensuring that the total numbers of data units in \textbf{Invisible} and \textbf{Visible} are the same and that the divergence of the distribution is less than the threshold $r=0.02$ (using the same metric as in \textit{systematicity}). We construct the test set using all samples with more than $N$ data units in the original test. We ensure that any data unit in the test set is present in both \textbf{Invisible} and \textbf{Visible}. Our experiments try the number threshold $N \in \{3, 4, 5\}$. Figure~\ref{fig:pro} shows an example of dataset construction.

\subsubsection{Dataset Statistics}
\par Samples in WebNLG with 6 and 7 data units only cover four domains: \textit{Astronaut}, \textit{Monument}, \textit{University}, and \textit{Company}. To avoid inconsistent domain distributions of training sets, we only use samples from these four domains to construct the datasets for the \textit{productivity} evaluation on WebNLG. Table~\ref{pro_info} shows the number of samples in training sets with each number of input triples. For $N\in\{3,4,5\}$, the test set of WebNLG contains 219 / 153 / 99 samples, and the test set of E2E contains 1,314 / 1,002 / 477 samples.

\subsubsection{Evaluation}
We train the model on \textbf{Invisible} and \textbf{Visible} respectively and test the performance of the two trained models on the test set. We evaluate \textit{productivity} of the model by the performance on \textbf{Invisible}. We use the performance on \textbf{Visible} as a bound to analyze the \textit{productivity} level of the model.

\subsubsection{Results and Analysis}

\par Table~\ref{pro_res} shows the results of the \textit{productivity} evaluation. On WebNLG, T5-11b performs best on \textbf{Invisible} with different thresholds. On E2E, the best performing model on \textbf{Invisible} with each threshold is one of the LLMs. The LLMs overall show stronger \textit{productivity} than the smaller LMs. However, all models, including LLMs, show performance gaps on \textbf{Invisible} and \textbf{Visible} on both WebNLG and E2E. As \textbf{Invisible} and \textbf{Visible} have the same total number of data units and close distribution of data units, the gaps are attributed to differences in the visibility of samples with the number of input data units exceeding the threshold, indicating that when the model cannot see samples with the number of input data units exceeding the threshold during training, it is unable to handle such samples as well as when it can see. This reflects a deficiency in \textit{productivity} of the model. The performance gaps of most models on \textbf{Invisible} and \textbf{Visible} are more significant for smaller thresholds, indicating that the deficiency in \textit{productivity} is more pronounced when the maximum number of input data units within a sample seen during training decreases. In conclusion, the LLMs overall show an improvement in \textit{productivity} compared to the smaller LMs but do not eliminate the deficiency in \textit{productivity} of the model.

\begin{table*}[h]

\centering
\resizebox{\textwidth}{!}{
\begin{tabular}{l|cccccc|cccccc}
\hline
            & \multicolumn{6}{c|}{WebNLG}                                                                                                                                    & \multicolumn{6}{c}{E2E}                                                                                                                                        \\ \hline
            & \multicolumn{2}{c|}{Fidelity}               & \multicolumn{2}{c|}{Ordering}               & \multicolumn{1}{c|}{\multirow{2}{*}{CWIO}} & \multirow{2}{*}{PERF} & \multicolumn{2}{c|}{Fidelity}               & \multicolumn{2}{c|}{Ordering}               & \multicolumn{1}{c|}{\multirow{2}{*}{CWIO}} & \multirow{2}{*}{PERF} \\ \cline{2-5} \cline{8-11}
            & PBH            & \multicolumn{1}{c|}{POH}   & PBH            & \multicolumn{1}{c|}{POH}   & \multicolumn{1}{c|}{}                      &                       & PBH            & \multicolumn{1}{c|}{POH}   & PBH            & \multicolumn{1}{c|}{POH}   & \multicolumn{1}{c|}{}                      &                       \\ \hline
T5-large    & 97.56          & \multicolumn{1}{c|}{1.67}  & 87.15          & \multicolumn{1}{c|}{6.84}  & \multicolumn{1}{c|}{+0.13}                 & 67.95                 & 91.39          & \multicolumn{1}{c|}{6.80}  & 77.22          & \multicolumn{1}{c|}{10.15} & \multicolumn{1}{c|}{+0.51}                 & 63.07                 \\
BART-large  & 97.65          & \multicolumn{1}{c|}{0.94}  & 88.69          & \multicolumn{1}{c|}{3.98}  & \multicolumn{1}{c|}{+0.10}                 & 66.96                 & 98.05          & \multicolumn{1}{c|}{0.90}  & 82.26          & \multicolumn{1}{c|}{3.59}  & \multicolumn{1}{c|}{+0.52}                 & 62.58                 \\
GPT-2-large & 90.55          & \multicolumn{1}{c|}{6.86}  & 82.64          & \multicolumn{1}{c|}{9.90}  & \multicolumn{1}{c|}{+0.11}                 & 67.64                 & 74.37          & \multicolumn{1}{c|}{19.22} & 68.08          & \multicolumn{1}{c|}{10.99} & \multicolumn{1}{c|}{+0.50}                 & 62.60                 \\
T5-11b      & \textbf{99.10} & \multicolumn{1}{c|}{0.64}  & \textbf{89.05} & \multicolumn{1}{c|}{4.53}  & \multicolumn{1}{c|}{+0.10}                 & 68.47                 & \textbf{99.12} & \multicolumn{1}{c|}{0.60}  & \textbf{82.75} & \multicolumn{1}{c|}{3.68}  & \multicolumn{1}{c|}{+0.57}                 & 62.56                 \\
Mistral-7b  & 96.49          & \multicolumn{1}{c|}{2.67}  & 86.29          & \multicolumn{1}{c|}{7.80}  & \multicolumn{1}{c|}{+0.11}                 & \textbf{68.69}        & 96.49          & \multicolumn{1}{c|}{3.08}  & 82.28          & \multicolumn{1}{c|}{4.46}  & \multicolumn{1}{c|}{+0.42}                 & \textbf{63.91}        \\
Llama-2-13b & 96.69          & \multicolumn{1}{c|}{2.33}  & 87.28          & \multicolumn{1}{c|}{6.86}  & \multicolumn{1}{c|}{+0.09}                 & 68.07                 & 96.88          & \multicolumn{1}{c|}{2.75}  & 78.50          & \multicolumn{1}{c|}{7.54}  & \multicolumn{1}{c|}{+0.46}                 & 62.81                 \\ \hline
T5-large    & 94.63          & \multicolumn{1}{c|}{4.55}  & 53.56          & \multicolumn{1}{c|}{39.98} & \multicolumn{1}{c|}{+0.81}                 & 65.53                 & 98.36          & \multicolumn{1}{c|}{1.62}  & 37.28          & \multicolumn{1}{c|}{43.95} & \multicolumn{1}{c|}{+0.95}                 & 55.74                 \\
BART-large  & 92.45          & \multicolumn{1}{c|}{5.94}  & 54.78          & \multicolumn{1}{c|}{38.14} & \multicolumn{1}{c|}{+0.76}                 & 64.12                 & 97.25          & \multicolumn{1}{c|}{2.67}  & 37.58          & \multicolumn{1}{c|}{43.91} & \multicolumn{1}{c|}{+0.95}                 & 56.98                 \\
GPT-2-large & 81.06          & \multicolumn{1}{c|}{15.80} & 54.84          & \multicolumn{1}{c|}{37.74} & \multicolumn{1}{c|}{+0.76}                 & 65.07                 & 85.34          & \multicolumn{1}{c|}{11.77} & \textbf{38.96} & \multicolumn{1}{c|}{42.31} & \multicolumn{1}{c|}{+0.95}                 & 56.52                 \\
T5-11b      & \textbf{96.58} & \multicolumn{1}{c|}{3.01}  & 54.93          & \multicolumn{1}{c|}{38.81} & \multicolumn{1}{c|}{+0.76}                 & 66.04                 & \textbf{99.28} & \multicolumn{1}{c|}{0.72}  & 37.05          & \multicolumn{1}{c|}{43.56} & \multicolumn{1}{c|}{+0.94}                 & 56.21                 \\
Mistral-7b  & 94.65          & \multicolumn{1}{c|}{4.64}  & \textbf{54.95} & \multicolumn{1}{c|}{38.46} & \multicolumn{1}{c|}{+0.79}                 & \textbf{66.29}        & 97.95          & \multicolumn{1}{c|}{1.93}  & 37.24          & \multicolumn{1}{c|}{43.40} & \multicolumn{1}{c|}{+0.94}                 & \textbf{57.37}        \\
Llama-2-13b & 91.44          & \multicolumn{1}{c|}{7.38}  & 54.59          & \multicolumn{1}{c|}{39.00} & \multicolumn{1}{c|}{+0.78}                 & 65.66                 & 97.97          & \multicolumn{1}{c|}{1.99}  & 37.38          & \multicolumn{1}{c|}{43.68} & \multicolumn{1}{c|}{+0.95}                 & 56.52                 \\ \hline
\end{tabular}
}

\caption{\label{ord_res}Results of models trained on \textbf{Original} (above) and \textbf{Match} (below) for the \textit{order invariance} evaluation. CWIO refers to the correlation with the input order. PERF refers to the performance on the original test set.}
\end{table*}
\subsection{Order Invariance}

The third aspect we evaluate is \textit{order invariance}. This notion is previously studied by \citet{DBLP:journals/corr/abs-2305-17926}, who finds that LLMs are sensitive to the order of options in multiple choice task. In the data-to-text generation task, \textit{order invariance} refers to the ability that a model's output text maintains the fidelity and proper ordering of data when the same unordered set of data is input in different orders. Having \textit{order invariance} means that the model can decompose the input into the set of data units and recombine them properly, regardless of the order of data units in the input, which reflects compositional generalization. In practical application scenarios, there are often cases where the data does not have a known linear order, and thus the model is required to have \textit{order invariance} to ensure the fidelity and proper data ordering of the output texts under any data input order.

\par In the \textit{order invariance} evaluation, for the same set of data units, we use two different input orders and then evaluate whether outputs maintain the fidelity and proper data ordering under both input orders. Further, we investigate the effect of the training process on \textit{order invariance}. We construct a training set in which data units are arranged in the input in the same order as they appear in the text. We evaluate whether using such a training set makes the model more inclined to arrange data units in the text according to input order and whether it affects the \textit{order invariance} of the model.

\subsubsection{Dataset Construction}
\label{sec:order_algorithm}
\par We design a search algorithm to find the occurrence position of data units in the text (see Appendix~\ref{app:ord_algo} for details). For each data-text pair in the original training set, we arrange the data units in the input according to their occurrence in the text, forming the training set \textbf{Match (M)}. Correspondingly, \textbf{Original (O)} refers to the original training set.

\begin{figure}
    \centering
    \includegraphics[width=\columnwidth]{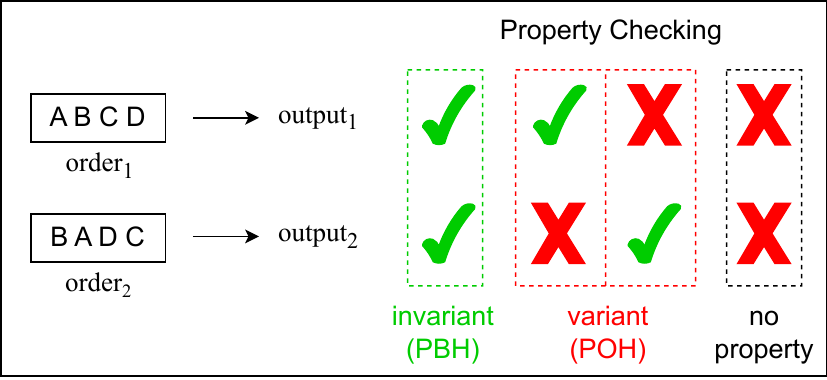}
    \caption{An illustration of the \textit{order invariance} evaluation. Each letter (A\textasciitilde D) denotes a data unit. For a certain property, the evaluation checks whether the output has that property. $\checkmark$ means yes and $\times$ means no.}
    \label{fig:ord}
\end{figure}

\subsubsection{Dataset Statistics}
\par For the \textit{order invariance} evaluation on fidelity and proper data ordering, we remove samples with only one data unit and samples where the order of the data units in the text cannot be determined. The test set of WebNLG contains 1,559 samples, and the test set of E2E contains 1,623 samples.

\subsubsection{Evaluation}
\par We train the model on \textbf{Original}. For each sample of the original test set, we randomize the order of the input data units to form two different inputs. We determine the set of data units contained in the output and the order of the data units, and then consider two properties: (1) The output is considered to have \textbf{fidelity} if the set of data units exactly matches the input. (2) The output is considered to have \textbf{proper data ordering} if the order of the data units satisfies $k>0$ with the order of at least one reference text, where $k\in[-1, 1]$ is the Kendall coefficient \cite{kendall}, which measures the correlation of two orders. For each of the two properties, we evaluate the proportion of both outputs having the property (PBH) and the proportion of only one output having the property (POH). A model with high \textit{order invariance} on the property should have a higher PBH. Relatively, POH reflects the order variance of the model. Figure~\ref{fig:ord} shows an illustration of the evaluation. 

\subsubsection{Additional Tests}
\par To investigate the effect of the order consistency of data units in input and output in the training set, we train the model on \textbf{Match} and perform additional tests. Besides fidelity and proper data ordering in the evaluation, we also perform the following tests on the models trained on \textbf{Original} and \textbf{Match}. First, for the input and model output of the original test set, we determine the order of data units in the output, and then calculate its correlation with the input order of the data units (CWIO). We use the Kendall coefficient to measure the correlation. A higher correlation means that the model is more inclined to arrange data units in the text according to input order. Second, we test the performance of the model on the original test set to see the effect of different training sets on the performance.

\subsubsection{Results and Analysis}

\par Table~\ref{ord_res} shows the results of the \textit{order invariance} evaluation. When trained on \textbf{Original}, on fidelity, T5-11b has the highest PBH on both WebNLG and E2E, showing the strongest \textit{order invariance}. As a smaller LM, BART-large has the second highest PBH, which is higher than LLMs Mistral-7b and Llama-2-13b. From the POH we can see that all models show \textit{order variant} cases on fidelity, i.e., for two input orders of the same set of data units, a model may show fidelity in one order but not in the other. On proper data ordering, the results are similar to fidelity and show a larger proportion of \textit{order variant} cases. This means that for two input orders of data units, the two outputs of the model may differ in their data ordering, where one is proper and the other is not. Overall, the models are deficient in \textit{order invariance} on both fidelity and proper data ordering. 
\par Compared to \textbf{Original}, when trained on \textbf{Match}, the CWIO of the model is significantly higher, indicating that the model is more inclined to arrange the data units in the text according to input order. This inclination about ordering leads to a decrease in \textit{order invariance} on proper data ordering. An unexpected finding is that the inclination also affects \textit{order invariance} on fidelity, overall leading to a decrease on WebNLG and an increase on E2E (see Appendix~\ref{app:qua_ord} for the discussion). The performance of the model trained on \textbf{Match} is significantly lower than on \textbf{Original}, indicating that high order consistency of data units in input and output during training negatively affects the performance when the order of input data units is arbitrary.

\subsection{Rule Learnability}
\par Models with high compositionality have the “willingness to prefer rules over memorization” \cite{PCFG}, i.e., they tend to apply observed rules to recombine elements rather than simply memorizing combinations of elements. Based on this understanding, we propose the last aspect of the evaluation, \textit{rule learnability}, which refers to the ability to learn rules from training and apply them during testing. Our evaluation focuses on the \textit{copy rule} \cite{DBLP:conf/inlg/GehrmannDER18} in data-to-text generation, which refers to the rule that certain information involved in the text (e.g., entities, numeric values) should be copied directly from the data to ensure the fidelity of the text. 
\par In the \textit{rule learnability} evaluation, we replace some entities or numeric values that should be copied with phrases that hide information, and then check whether the model correctly applies the \textit{copy rule}. A correct copy should not have \textit{omissions} of phrases that hide information or \textit{hallucinations} of outputting entities and numeric values that have been hidden. If the model only memorizes specific mappings that conform to the \textit{copy rule} during training, rather than actually learning the \textit{copy rule}, then it will not be able to correctly apply the \textit{copy rule} to the phrases that hide information.

\begin{figure}
    \centering
    \includegraphics[width=\columnwidth]{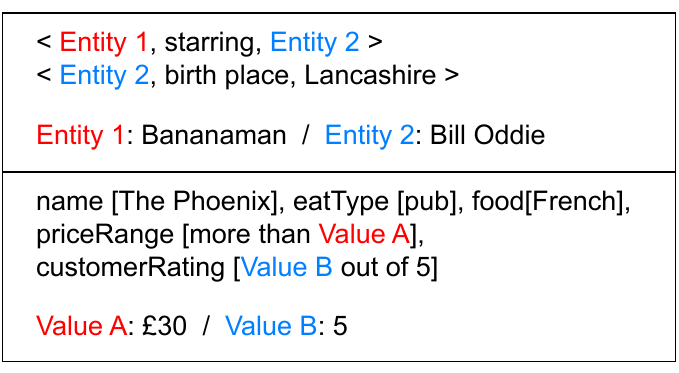}
    \caption{An example of dataset construction for the \textit{rule learnability} evaluation.}
    \label{fig:reg}
\end{figure}

\subsubsection{Dataset Construction}
\par On WebNLG, the \textit{copy rule} is mainly applied to entities. For each sample in the original WebNLG test set, we find the entities that act as subjects and are copied in every reference text, and replace these entities in the input with "Entity $i$" ($i$ denotes the entity's label, which is used to distinguish between different entities). On E2E, the \textit{copy rule} is mainly applied to values, and we focus on numeric values. Similar to WebNLG, we replace the numeric value with "Value $i$". If a value contains more than one numeric value, only the first one will be replaced. Figure~\ref{fig:reg} shows an example of dataset construction.

\subsubsection{Dataset Statistics}
\par For the \textit{rule learnability} evaluation, on WebNLG, we retain only samples in which there is at least one entity that satisfies the replacement condition. The final test set contains 1,614 samples. On E2E, since the training data guarantees copies of values, we can construct samples without reference texts to cover more combinations. We enumerate the values of 6 attributes (except the attribute \textit{near}, which is similar to \textit{name}) and ensure that at least one value contains the numeric value, resulting in 1,440 samples in the final test set.

\subsubsection{Evaluation}
We train the model on the original training set and then check the output of the model on the replaced inputs. The result of checking each sample can be represented as \textbf{(a, b)}, where \textbf{a} $\in$ \textbf{\{0, 1\}} indicates whether all phrases that hide information are copied correctly (using fuzzy matching, see Appendix~\ref{app:reg_match} for details), and \textbf{b} $\in$ \textbf{\{0, 1\}} indicates whether the hidden entities or numeric values appear. In E2E, for a hidden value, we also consider \textbf{b = 1} if other possible values corresponding to its attribute appear. In the representation of the result, \textbf{a = 0} implies \textit{omissions} and \textbf{b = 1} implies \textit{hallucinations}. Of the four possible results, only \textbf{(1, 0)} indicates that the \textit{copy rule} is correctly applied. We count the proportions of the four cases and evaluate the \textit{rule learnability} by the proportion of \textbf{(1, 0)}.

\subsubsection{Results and Analysis}
\par Table~\ref{reg_res} shows the results of \textit{rule learnability} evaluation. On WebNLG, all models apply the \textit{copy rule} less than 90\% correctly. The errors are mainly concentrated on the \textbf{(0, 0)} case. This case indicates that the model does not have the \textit{hallucinations} of outputting entities that have been hidden, but it has \textit{omissions} of phrases that hide information. Among all the models, T5-large and BART-large have relatively high correct rates. The LLMs do not show higher correct rates compared to the smaller LMs. All LLMs have a correct rate of less than 80\%.

\par The results shown on E2E are different. On E2E, the LLMs have high correct rates and outperform the smaller LMs. Among the LLMs, both Mistral-7b and Llama-2-13b are almost completely correct. Among the smaller LMs, BART-large and GPT-2-large show very low correct rates. Their proportions of \textbf{(0, 1)} are both high, indicating that there are serious \textit{hallucinations} of outputting numeric values that have been hidden. When outputting these numeric values, the model tends not to output the corresponding phrases that information, resulting in \textit{omissions}. Their proportions of \textbf{(0, 0)} also indicate the presence of simple \textit{omissions} unrelated to the \textit{hallucinations}.
\par In summary, the results show that all models, including LLMs, are unable to achieve high correct copy rates on both WebNLG and E2E, and that \textit{omissions} and \textit{hallucinations} are prevalent in the models. This indicates that for \textit{copy rules} in data-to-text generation, the models are deficient in \textit{rule learnability} and need further improvement.

\begin{table}
\centering
\begin{tabular}{l|cccc}
\hline
            & \textbf{(0, 0)} & \textbf{(0, 1)} & \textbf{(1, 0)} & \textbf{(1, 1)} \\ \hline
T5-large    & 10.16           & 0.31            & \textbf{89.32}  & 0.21            \\
BART-large  & 10.64           & 1.30            & 87.59           & 0.48            \\
GPT-2-large & 19.43           & 1.69            & 78.44           & 0.43            \\
T5-11b      & 17.35           & 3.02            & 79.62           & 0.02            \\
Mistral-7b  & 19.08           & 1.40            & 79.04           & 0.48            \\
Llama-2-13b & 21.15           & 0.45            & 78.11           & 0.29            \\ \hline
T5-large    & 2.64            & 1.44            & 95.93           & 0.00            \\
BART-large  & 13.17           & 57.57           & 29.26           & 0.00            \\
GPT-2-large & 15.28           & 48.19           & 36.06           & 0.46            \\
T5-11b      & 0.05            & 2.38            & 97.57           & 0.00            \\
Mistral-7b  & 0.65            & 0.00            & \textbf{99.35}  & 0.00            \\
Llama-2-13b & 0.86            & 0.00            & 99.14           & 0.00            \\ \hline
\end{tabular}

\caption{\label{reg_res}Results of the \textit{rule learnability} evaluation on WebNLG (above) and E2E (below). Each column represents the proportion of the corresponding case.}
\end{table}

\section{Conclusions}

In this work, we propose \textsc{SPOR}, a comprehensive and practical evaluation method for compositional generalization in data-to-text generation, which includes four aspects of manifestations: \textit{systematicity}, \textit{productivity}, \textit{order invariance}, and \textit{rule learnability}. We demonstrate on WebNLG and E2E how \textsc{SPOR} enables evaluations without additional manual annotations based on existing datasets. We evaluate some existing language models, including LLMs. We find that the models are deficient in various aspects of compositional generalization in data-to-text generation and need further improvement. Our work supports comprehensive research on different manifestations of compositional generalization in data-to-text generation and provides a framework for identifying and evaluating improvements in this ability of language models.

\section*{Limitations}
A limitation of our work is the limited size of the models evaluated. Although we include some LLMs in our evaluation, due to the need for fine-tuning with limited resources, the size of the LLMs does not exceed 13b. Resource constraints make it difficult to apply fine-tuning methods on larger LMs, and there is currently no effective method for directly applying larger LMs to data-to-text generation. One possible method is in-context learning, which performs inference directly but adds a prefix to the input that demonstrates a small number of samples for the model to learn. In the in-context learning style, the training phase of compositional generalization corresponds to the sample demonstration in the prefix, and the evaluation needs to consider the method of sample demonstration selection. We will continue to follow the progress of applying larger LMs to data-to-text generation and explore evaluation methods for compositional generalization in data-to-text generation of larger LMs.
\section*{Ethics Statement}
\par The datasets and models we use are open-source and we use them for scientific research purposes only. The datasets we construct will also be open source for scientific research purposes. The datasets we use and construct do not contain any information that names or uniquely identifies individual people or offensive content.
\par Since we use the realistic dataset WebNLG, we are particularly concerned with data faithfulness, i.e., all data in the reconstructed evaluation dataset must not show information that contradicts the original realistic dataset, which may be inconsistent with the real world and may be harmful. In the \textit{systematicity}, \textit{productivity}, and \textit{order invariance} evaluations, we do not modify the information in any triple. In the \textit{rule learnability} evaluation, we only hide the information, and no new information is generated. Therefore, the data used in the evaluation do not contain information that contradicts the original realistic dataset.
\par The AI assistant we use in our work is Copilot (for simple code completion).
\section*{Acknowledgements}

This work was supported by National Science and Technology Major Project (2022ZD0116308). The corresponding author is Houfeng Wang.
\par We would like to thank the anonymous reviewers for their recognition and valuable suggestions for our work. These suggestions helped us to revise the work to make it more solid.

\bibliography{anthology,custom}

\clearpage
\appendix

\section{Model Details}
\label{app:model}
\par The models we evaluate include T5-large (738M), BART-large (406M), GPT2-large (774M), T5-11b, Mistral-7b, and Llama-2-13b. All models are downloaded from HuggingFace, and training and inference are based on the transformers library. Each item in our experiment is done on a single NVIDIA A800 80G GPU.
\par For model input, we use the linearization method \cite{investigating, T5forD2T}. For WebNLG, we add the special identifiers <head>, <relation>, and <tail> before the subject, predicate, and object of each triple, and then linearly concatenate all triples to form the input. For E2E, We form the input by linearly concatenating each attribute-value pair in the form of "attribute[value]". Following \citet{investigating}, for WebNLG, we add a prefix “translate from Triple to Text:” before the input. Similarly, we use the prefix "translate from MR to Text:" for E2E.
\par For \textit{systematicity} and \textit{productivity} evaluations, we report the best results on the test set among all checkpoints. For \textit{order invariance} and \textit{rule learnability} evaluations, we report the results of the checkpoint that has the best performance on the original test set.

\section{Qualitative Analysis of Evaluations}
\label{app:qua}
Table~\ref{tab_qua_1} \textasciitilde~\ref{tab_qua_3} show some specific samples with model outputs in each aspect of the evaluation.

\subsection{Systematicity}
Table~\ref{tab_qua_1} shows samples from Llama-2-13b in the \textit{systematicity} evaluation. On WebNLG, the issue on fidelity is the omission of data units, and the issue on fluency is the stiff expression (the model repeatedly enumerates data units by applying the same pattern, and lacks fluency in articulation). On E2E, the issues center on fluency similar to those shown on WebNLG. The stiff expression can be attributed to the difficulty of models trained on the \textbf{Atom} in handling unseen combinations.

\subsection{Productivity}
Table~\ref{tab_qua_1b} shows samples from Llama-2-13b in the \textit{productivity} evaluation. The issues center on fidelity. In addition to the omissions present on WebNLG and E2E, hallucinations are found on E2E. The fidelity issue can be attributed to the difficulty of models trained on \textbf{Invisible} in handling a larger number of input data units.

\subsection{Order Invariance}
\label{app:qua_ord}
Table~\ref{tab_qua_2} shows samples from Llama-2-13b in the \textit{order invariance} evaluation. On WebNLG, for the model trained on \textbf{Original}, both outputs have fidelity. However, the data ordering of Output 1 is improper, while that of Output 2 is proper (for < Trance music, stylistic origin, Pop music >, it should be next to < Andrew Rayel, genre, Trance music >, not isolated at the end). For the model trained on \textbf{Match}, the order of the output data units is consistent with the input order. When the input order is not the proper data ordering, the model may try to apply complex grammar on an unnatural order of data units, which results in some data units not being generated as demonstrated in the sample. On E2E, the two outputs on \textbf{Original} are consistent in ordering but vary in fidelity. The two outputs on \textbf{Match} have exactly the same data ordering as the inputs, resulting in a stiff expression. However, from the experimental results, such a form of output on improves \textit{order invariance} on fidelity on E2E. We hypothesize that due to the relatively simple grammar of E2E, this form does not lead to omissions as on WebNLG, and the model may be easier to maintain fidelity because there is no need to rearrange the data units.

\subsection{Rule Learnability}
\par Table~\ref{tab_qua_3} shows samples of error cases in the \textit{rule learnability} evaluation. The most frequent error case on WebNLG is \textbf{(0, 0)}. In the sample of \textbf{(0, 0)} on WebNLG, there is no hallucination in the output but "Entity 1" is omitted, resulting in a factual error. The other two samples demonstrate cases with hallucinations.  On a realistic dataset like WebNLG, the hallucination may be a correct inference based on known information but does not satisfy the requirement for fidelity in data-to-text generation. The poorer performing models on E2E, such as BART-large / GPT-2-large, have a large proportion of \textbf{(0, 1)} cases. In the sample of \textbf{(0, 1)} on E2E, the model outputs "5 out of 5" instead of "Value B of 5", which is a hallucination with the omission. On E2E, known information is irrelevant to the hidden numeric value, so the hallucination is unfounded. The sample of \textbf{(0, 0)} demonstrates an omission unrelated to the hallucination, which is the only case of errors for the better performing models on E2E such as Mistral-7b / Llama-2-13b.

\section{Search Algorithm for Order-Invariance Evaluation}
\label{app:ord_algo}

\par For each data-text pair in WebNLG, we first locate where the entities in the data appear in the text. Although most of the entities appear unchanged in the text, variations still exist, such as token discontinuities or token distortions. However, discontinuous tokens are not too far away from each other, and the degree of token distortion is not too large. Therefore, we use the following algorithm for localization:

\par 1. We first slice the entity into tokens, and for each token $t$, find the set of candidate-matching tokens in the text with the smallest edit distance from $t$ and no more than min ($2$, length of $t$).
\par 2. Keep all non-empty candidate sets, and then use depth-first search to select a position in each candidate set such that the final variance of all positions is minimized as the token position representation of the entity. If there are multiple minimum variance representations, then all are retained.
\par 3. The entities are sorted by the number of position representations retained from smallest to largest, and then one representation is selected for each entity and the smallest position number in the representation is used to represent that entity. We require that the position number representing an entity cannot appear in the representations of other entities, and if it cannot be satisfied, then the position number of this entity is set to a large boundary value (the percentage of such cases is about 1.6\%).

\par After determining the position number of each entity, we determine the order of triples. We consider the set of triples as an undirected graph, and each triple represents a connected edge between the subject and the object. For each triple, if the degree of the subject and object are different, we take the position of the entity with the smaller degree to represent the position of the triple, otherwise, we take the larger of the two entity positions to represent the position of the triple. According to the position number of triple, we get the order of triple. The order relationship between triples with the same position number follows the input.

\par On E2E, since the training data guarantees copies of values, we use strict matching to localize the values.

\section{Fuzzy Matching for Rule-Learnability Evaluation}
\label{app:reg_match}
\par In the \textit{rule learnabilit}y evaluation, for the checking of copying phrases that hide information, we find that there are cases where the model does not perform strict copying, but semantically completes the copying, which should also be considered correct. Therefore, in addition to strictly correct copying, the following cases are also considered as correct copying:
\begin{itemize}[itemsep=2pt,topsep=2pt,parsep=2pt]

\item Case is ignored. For example, "entity 1" and "value b" are considered correct.
\item Numeric symbols can be changed to ordinal numbers. For example, "1st Entity" is considered correct.
\item If the symbol is copied, it is allowed not to copy "Entity" or "Value". For example, "Its customer rating is B out of 5." is considered correct.
\end{itemize}
\par The fuzzy matching covers most cases of semantically completed copies, which makes the checking of copying more accurate.

\begin{table*}

\centering

\begin{tabular}{p{0.1\textwidth}|p{0.85\textwidth}}
\hline
\textbf{Dataset} & \textbf{Sample} \\
\hline
WebNLG &  
{
    \par \textbf{Input:} < Ayam penyet, region, Malaysia >, < Ayam penyet, country, Java >, < Ayam penyet, ingredient, Fried chicken >, < Ayam penyet, main ingredient, Squeezed or smashed fried chicken served with sambal >, < Ayam penyet, serving temperature, Hot >
    \par \textbf{Combination:} Ayam penyet is a dish from Malaysia and Java. It includes fried chicken which is squeezed or smashed and served with sambal. It should be served hot.
    \par \textbf{Atom:} Ayam penyet is a dish from the region of Malaysia and Java. It contains fried chicken and is served hot.
    \par \textbf{Performance:} (73.90, 56.98) \ \ \textbf{Issue:} Omission of data units
} \\
\cline{2-2}
& {
    \par \textbf{Input:} < Spain, leader, Felipe VI of Spain >, < Spain, language, Spanish language >, < Spain, currency, Euro >, < Ajoblanco, country, Spain >, < Spain, demonym, Spaniards >
    \par \textbf{Combination:} Ajoblanco is a dish from Spain, where the currency is the euro and the language is Spanish. The country is led by Felipe VI and the people who live there are called Spaniards.
    \par \textbf{Atom:} Ajoblanco is a food found in Spain, where Felipe VI of Spain is the leader, Spanish is spoken, the Euro is the currency and Spaniards live.
    \par \textbf{Performance:} (62.93, 8.43) \ \ \textbf{Issue:} Stiff expression
}\\
\hline
E2E & {
    \par \textbf{Input:} name[Wildwood], eat type[restaurant], food[French], area[riverside], near[Raja Indian Cuisine]
    \par \textbf{Combination:} Wildwood is a French restaurant near Raja Indian Cuisine in the riverside area.
    \par \textbf{Atom:} Wildwood is a restaurant providing French food It is located in the riverside. It is near Raja Indian Cuisine.
    \par \textbf{Performance:} (59.89, 24.93) \ \ \textbf{Issue:} Stiff expression / Missing punctuation
}
\\ \hline
\end{tabular}

\caption{\label{tab_qua_1} Samples from Llama-2-13b in the \textit{systematicity} evaluation. \textbf{Combination} and \textbf{Atom} refer to the output of the same model trained on the corresponding dataset. \textbf{Performance} refers to the performance scores of the two outputs.}
\end{table*}
\begin{table*}

\centering

\begin{tabular}{p{0.1\textwidth}|p{0.85\textwidth}}
\hline
\textbf{Dataset} & \textbf{Sample} \\
\hline
WebNLG &  
{
    \par \textbf{Input:} < Elliot See, alma mater, University of Texas at Austin >, < University of Texas at Austin, affiliation, University of Texas System >, < Elliot See, birth place, Dallas >, < Elliot See, death place, St. Louis >, < Elliot See, status, Deceased >
    \par \textbf{Visible:} Elliot See was born in Dallas and died in St. Louis. He attended the University of Texas at Austin, which is affiliated to the University of Texas system.
    \par \textbf{Invisible:} Elliot See was born in Dallas, Texas, graduated from the University of Texas at Austin and died in St Louis.
    \par \textbf{Performance:} (82.85, 70.91) \ \ \textbf{Issue:} Omission of data units
} \\
\hline
E2E & {
    \par \textbf{Input:} name[Browns Cambridge], eat type[coffee shop], food[Chinese], customer rating[average], area[city centre], area[riverside], family friendly[no], near[Crowne Plaza Hotel]
    \par \textbf{Visible:} Browns Cambridge is a Chinese coffee shop located in the city centre near the Crowne Plaza Hotel. It is not family friendly and has an average customer rating. It is located in the riverside area.
    \par \textbf{Invisible:} Browns Cambridge is a coffee shop providing Chinese food It is located in the city centre. Its customer rating is average. It is not family friendly. It is near Crowne Plaza Hotel.
    \par \textbf{Performance:} (62.84, 53.84) \ \ \textbf{Issue:} Omission of data units 
}
\\ 
\cline{2-2}
 & {
    \par \textbf{Input:} name[Clowns], eat type[coffee shop], food[English], customer rating[5 out of 5], near[Clare Hall]
    \par \textbf{Visible:} Clowns is a coffee shop near Clare Hall. It serves English food and has a customer rating of 5 out of 5.
    \par \textbf{Invisible:} Clowns is a coffee shop providing English food in the higher price range. Its customer rating is 5 out of 5. It is near Clare Hall.
    \par \textbf{Performance:} (64.59, 23.47) \ \ \textbf{Issue:} Hallucination
}
\\ \hline
\end{tabular}

\caption{\label{tab_qua_1b} Samples from Llama-2-13b in the \textit{productivity} evaluation with the threshold $N=3$. \textbf{Visible} and \textbf{Invisible} refer to the output of the same model trained on the corresponding dataset. \textbf{Performance} refers to the performance scores of the two outputs.}
\end{table*}
\begin{table*}

\centering

\begin{tabular}{p{0.1\textwidth}|p{0.85\textwidth}}
\hline
\textbf{Dataset} & \textbf{Sample} \\
\hline
WebNLG &  
{
    \par \textbf{Input:} < Trance music, stylistic origin, Pop music >, < Andrew Rayel, genre, Trance music >, < Jwaydan Moyine, associated band/associated musical artist, John Digweed >, < Andrew Rayel, associated band/associated musical artist, Jwaydan Moyine >
} \\ \cline{2-2} 
 & {
    \par \textbf{Input Order 1:} < Trance music, stylistic origin, Pop music > → < Andrew Rayel, genre, Trance music > → < Jwaydan Moyine, associated band/associated musical artist, John Digweed > → < Andrew Rayel, associated band/associated musical artist, Jwaydan Moyine >
    \par \textbf{Original 1:} \textcolor{red}{Andrew Rayel} is a \textcolor{red}{Trance} musician who is associated with the musical artist \textcolor{red}{Jwaydan Moyine}. Moyine is associated with the musical artist \textcolor{red}{John Digweed}. Trance music originated from \textcolor{red}{pop music}.
    \par \textbf{Match 1:} \textcolor{red}{Trance music} originated from \textcolor{red}{pop music} and is performed by artists such as \textcolor{red}{Andrew Rayel} and \textcolor{red}{John Digweed}.
} \\ \cline{2-2}
  &  {
  \par \textbf{Input Order 2:} < Jwaydan Moyine, associated band/associated musical artist, John Digweed > → < Trance music, stylistic origin, Pop music > → < Andrew Rayel, associated band/associated musical artist, Jwaydan Moyine > → < Andrew Rayel, genre, Trance music >
    \par \textbf{Original 2:} \textcolor{red}{Andrew Rayel's} musical genre is \textcolor{red}{Trance} which has its origins in \textcolor{red}{Pop music}. He is associated with the musical artist \textcolor{red}{Jwaydan Moyine} who is associated with the musical artist \textcolor{red}{John Digweed}.
    \par \textbf{Match 2:} \textcolor{red}{Jwaydan Moyine} is associated with the musical artist \textcolor{red}{John Digweed} and with \textcolor{red}{Trance} musician \textcolor{red}{Andrew Rayel}.
    }
\\ \hline


E2E &  
{
    \par \textbf{Input:} name[The Eagle], family friendly[no], price range[cheap], area[city centre], near[Burger King], customer rating[average], food[Chinese], eat type[coffee shop]
} \\ \cline{2-2}
 & {
    \par \textbf{Input Order 1:} family friendly[no] → price range[cheap] → area[city centre]  → near[Burger King] → customer rating[average] → food[Chinese] → eat type[coffee shop]
    \par \textbf{Original 1:} The Eagle is a \textcolor{red}{cheap} \textcolor{red}{Chinese} \textcolor{red}{coffee shop} in the \textcolor{red}{city centre} near \textcolor{red}{Burger King}. It has an \textcolor{red}{average} customer rating.
    \par \textbf{Match 1:} The Eagle is \textcolor{red}{not family friendly}, but is \textcolor{red}{cheap}. It is located in the \textcolor{red}{city centre} near \textcolor{red}{Burger King}. It has an \textcolor{red}{average} customer rating and serves \textcolor{red}{Chinese} food. It is a \textcolor{red}{coffee shop}.
} \\ \cline{2-2}
  &  {
  \par \textbf{Order 2:} food[Chinese] → customer rating[average] → near[Burger King] → price range[cheap] → area[city centre] → family friendly[no] → eat type[coffee shop] 
    \par \textbf{Original 2:} The Eagle is a \textcolor{red}{cheap} \textcolor{red}{Chinese} \textcolor{red}{coffee shop} in the \textcolor{red}{city centre} near \textcolor{red}{Burger King}. It has an \textcolor{red}{average} customer rating and is \textcolor{red}{not family friendly}.
    \par \textbf{Match 2:} The Eagle serves \textcolor{red}{Chinese} food. It has an \textcolor{red}{average} customer rating, is near \textcolor{red}{Burger King}, is \textcolor{red}{cheap}, is in the \textcolor{red}{city centre}, is \textcolor{red}{not family friendly}, and is a \textcolor{red}{coffee shop}.
    }
\\ \hline
\end{tabular}

\caption{\label{tab_qua_2}Samples from Llama-2-13b in the \textit{order invariance} evaluation. \textbf{Original} and \textbf{Match} refer to the output of the same model trained on the corresponding dataset. We mark the position of data units in the output in \textcolor{red}{red}.}
\end{table*}
\begin{table*}

\centering

\begin{tabular}{p{0.1\textwidth}|p{0.1\textwidth}|p{0.71\textwidth}}
\hline
\textbf{Dataset} & \textbf{Case} & \textbf{Sample} \\
\hline
WebNLG &  \textbf{(0, 0)} & 
{
    \par \textbf{Input:} < Antares (rocket), manufacturer, Yuzhnoye Design Office >, < Entity 1, launch site, Vandenberg Air Force Base >, < Antares (rocket), comparable, Entity 1 >, <Antares (rocket), launch site, Mid-Atlantic Regional Spaceport >, < Entity 1, country origin, United States >
    \par \textbf{Hidden:} (Entity 1, Delta II)
    \par \textbf{Output:} The Antares rocket, manufactured by the Yuzhnoye Design Office, was launched from the Mid-Atlantic Regional Spaceport and the Vandenberg Air Force Base in the United States.
} \\ \cline{2-3} & \textbf{(0, 1)}
 & {
    \par \textbf{Input:} < Entity 1, academic discipline, Mathematics >, < Entity 1, first publication year, 1878 >, < Entity 1, abbreviation, Am. J. Math >, < Entity 1, issn number, 1080-6377 >
    \par \textbf{Hidden:} (Entity 1, American Journal of Mathematics)
    \par \textbf{Output:} \textcolor{red}{The American Journal of Mathematics} (abbreviated to Am. J. Math.) has the ISSN number 1080-6377.
} \\ \cline{2-3} & \textbf{(1, 1)}
 & {
    \par \textbf{Input:} < Entity 2, leader, Mulatu Teshome >, < Entity 2, leader, Hailemariam Desalegn >, < Addis Ababa, is part of, Addis Ababa Stadium >, < Entity 1, location, Addis Ababa >, < Addis Ababa, country, Entity 2 >
    \par \textbf{Hidden:} (Entity 1, Addis Ababa City Hall), (Entity 2, Ethiopia)
    \par \textbf{Output:} Addis Ababa Stadium is located in Addis Ababa, \textcolor{red}{Ethiopia}. \textcolor{blue}{Entity 1} is located in Addis Ababa. Mulatu Teshome and Hailemariam Desalegn are leaders of \textcolor{blue}{Entity 2}.
    }
\\ \hline


E2E &  \textbf{(0, 0)} & 
{
    \par \textbf{Input:} name[The Twenty Two], price range[less than Value A], area[city centre], eat type[pub], food[Italian], near[Café Rouge], family friendly[no], customer rating[Value B out of 5]
    \par \textbf{Hidden:} (Value A, £20), (Value B, 1/3/5)
    \par \textbf{Output:} The Twenty Two is a pub located in the city centre near Café Rouge. It serves Italian food and has a customer rating of \textcolor{blue}{Value B out of 5}. It is not family friendly.
} \\ \cline{2-3} & \textbf{(0, 1)}
 & {
    \par \textbf{Input:} name[Loch Fyne], eat type[coffee shop], customer rating[Value B out of 5], area[city centre], price range[more than Value A], food[English], near[The Sorrento]
    \par \textbf{Hidden:} (Value A, £30), (Value B, 1/3/5)
    \par \textbf{Output:} Loch Fyne is a coffee shop near The Sorrento in the city centre. It has a customer rating of \textcolor{red}{5 out of 5} and serves English food at a price range of \textcolor{blue}{more than Value A}.
} \\ \cline{2-3} & \textbf{(1, 1)}
 & {
    \par \textbf{Input:} name[Alimentum], area[city centre], near[The Six Bells], eat type[pub], food[Chinese], price range[more than Value A]
    \par \textbf{Hidden:} (Value A, £30)
    \par \textbf{Output:} \textcolor{blue}{more than Value A}, Alimentum is a pub that provides Chinese food in the \textcolor{red}{more than £30} price range. It is located in the city centre.
    }
\\ \hline
\end{tabular}

\caption{\label{tab_qua_3} Samples from Llama-2-13b / T5-11b / GPT-2-large that performs incorrect copies in the \textit{rule learnability} evaluation. \textbf{Hidden} indicates the entities or numeric values that are hidden (this part does not appear in inputs). We mark copies of phrases that hide information in \textcolor{blue}{blue} and occurrences of hidden entities or numerical values in \textcolor{red}{red}.}
\end{table*}

\end{document}